\documentclass[preprints,article,accept,pdftex]{mdpi} 

\firstpage{779}
\pubvolume{4}
\issuenum{3}
\articlenumber{38}
\pubyear{2022}
\copyrightyear{2022}
\datereceived{6 August 2022}
\dateaccepted{7 September 2022 }
\datepublished{10 September 2022}
\hreflink{https://\linebreak doi.org/10.3390/make4030038} 
\doinum{10.3390/make4030038}

\usepackage{nameref}

\newcommand{\bal}[1]{\overline{\mathcal{#1}}}
\renewcommand{\hl}[1]{#1}

\Title{Factorizable Joint Shift in Multinomial Classification}

\TitleCitation{Factorizable Joint Shift in Multinomial Classification}

\Author{\hl{Dirk Tasche}}

\AuthorNames{Dirk Tasche}

\AuthorCitation{Tasche, D.}

\address[1]{\hl{Independent Researcher, 8032 Zurich, Switzerland; dirk.tasche@gmx.net}}

\abstract{%
Factorizable joint shift (FJS) was recently proposed as a type of dataset shift for which the complete characteristics can be estimated from feature data observations on the test dataset by a method called Joint Importance Aligning.
For the multinomial (multiclass) classification setting, we derive a representation of factorizable joint shift in terms of the
source (training) distribution, the target (test) prior class probabilities and the target marginal distribution of the features.
On the basis of this result, we propose alternatives to joint importance aligning and, at the same time, point out that factorizable joint shift is not fully identifiable if no class label information on the test dataset is available and no additional assumptions are made.  Other results of the paper include correction formulae for the posterior class probabilities both under general dataset shift and factorizable joint shift. In addition, we investigate the consequences of assuming factorizable joint shift for the bias caused by sample selection.}


\begin{document}


\section{Introduction}

In machine learning terminology, dataset shift refers to the phenomenon that the joint distribution of features and labels on the training dataset used for learning a model may differ from the related joint distribution on the test dataset to which the model is going to be applied; see Storkey \cite{storkey2009training} or Moreno-Torres et al.~\cite{MorenoTorres2012521} for surveys and background information on dataset shift. Dataset shift can be the consequence of very different causes. For that reason, a catch-all treatment of general dataset shift is difficult if not impossible. As a workaround a number of specific types of dataset shift have been defined in order to introduce additional assumptions that allow for different tailor-made approaches to deal with the problem. The most familiar subtypes of dataset shift are prior probability shift and covariate shift, but more types are introduced on a continuing basis as there is a practice-driven need to do so.

Typically, under dataset shift, the test dataset observations of features are available, but the class labels cannot be observed. In this situation, it is impossible to know ex ante if covariate shift or prior probability shift (or something in between) has occurred.
However, estimates of models under assumptions of covariate shift and prior probability shift, respectively, tend to  differ conspicuously. As a consequence, additional assumptions need to be made in order to be able to choose between modelling options related to covariate shift and prior probability shift. Such additional assumptions may be phrased in terms of causality (Storkey
\cite{storkey2009training}): if the features can be considered ``causing'' the class labels, then models designed to deal with covariate shift are appropriate. Otherwise, if the class ``causes'' features, models targeting prior probability shift should be preferred.

He et al.~\cite{he2022domain} recently proposed ``factorizable joint shift'' (FJS) which generalises both prior probability shift and covariate shift. They went on with presenting the ``joint importance aligning'' method for estimating the characteristics of this type of shift. At first glance, He et al.\ hence seemed to provide a way to avoid choosing ex ante between covariate shift and prior probability shift models. Instead, ``joint importance aligning'' (plus some regularisation) appeared to be a method that functioned as a covariate shift model, prior probability shift model, or combined covariate and label shift model, as required by the characteristics of the test dataset.

By a detailed analysis of factorizable joint shift in multinomial classification settings, in this paper we point out that general factorizable joint shift is not fully identifiable if no class label information on the test dataset is available and no additional assumptions are made.
This is in contrast to the situations with covariate shift or prior probability shift. Therefore, circumspection is recommended with regard to potential deployment of ``joint importance aligning'' as proposed by He et al.~\cite{he2022domain}.

He et al.\ characterised factorizable joint shift by claiming that ``the biases coming from the data and the label are statistically independent''. This description might not fully hit the mark.
As we demonstrate in this paper, factorizable joint shift has little to do with statistical independence but should rather be interpreted as a structural property similar to the ``separation of variables'' which plays an important role for finding closed-form solutions to differential equations. We also argue that, in probabilistic terms, factorizable joint shift perhaps is better described as ``scaled density ratios'' shift.

The plan of this paper and its main research contributions are as follows:
\begin{itemize}
\item Section~\ref{se:setting} ``Setting the scene'' presents the assumptions, concepts and notation for the multinomial (or multiclass) classification setting of this paper.
\item Section~\ref{se:normal} ``General dataset shift in multinomial classification'' introduces a normal form for the joint density of features and class labels (Theorem~\ref{th:dens}) and derives in Corollary~\ref{co:correct} a generalisation of the correction formula for class posterior probabilities of \mbox{Saerens et al.~\cite{saerens2002adjusting}} and Elkan \cite{Elkan01}.
\item Section~\ref{se:factor} ``Factorizable joint shift'' defines this kind of dataset shift in a mathematically rigorous manner and presents a full representation in terms of the source (training) distribution, the target (test) prior class probabilities and the target marginal distribution of the features (Theorem~\ref{th:factorized}). In addition, a specific version of the posterior correction formula is given (Corollary~\ref{co:factorized}), and the description of factorizable joint shift as ``scaled density ratios'' shift is motivated.
Moreover, alternatives to the ``joint importance aligning'' of He et al.~\cite{he2022domain} are proposed (Section~\ref{se:alternative}).
\item Section~\ref{se:examples} ``Common types of dataset shift'' examines in a mathematically rigorous manner for a number of types of dataset shift mentioned in the literature if they are implied by or imply factorizable joint shift. The types of dataset shift treated in this section are prior probability shift, covariate shift, covariate shift with posterior drift, domain invariance and generalised label shift.
In addition, the posterior correction formulae specific for these types of dataset shift are presented.
\item Section~\ref{se:sample} ``Sample selection bias'' revisits the topic of dataset shift caused by sample selection bias and looks at the question of how the class-wise selection probabilities look like if the induced dataset shift is factorizable joint shift (Theorem~\ref{th:FJS.sel}).
\item Section~\ref{se:concl} ``Conclusions'' provides a short discussion of the important findings of the paper and points to some open research questions.
\end{itemize}


\section{Setting the Scene}
\label{se:setting}

In this paper, we use the following population-level description of the multinomial classification problem under dataset shift in terms of measure theory. See standard textbooks on probability theory like Billingsley \cite{billingsley1986probability} or
Klenke \cite{klenke2013probability} for formal definitions and background of the notions introduced in Assumption~\ref{as:setting}. See
Tasche \cite{tasche2022class} for a detailed reconciliation of the setting of this paper with the concepts and notation used in the mainstream machine learning literature.
\begin{Assumption}\label{as:setting}
$(\Omega, \mathcal{F})$ is a measurable space. The \emph{source distribution}
$P$ and the \emph{target distribution} $Q$ are probability measures on $(\Omega, \mathcal{F})$.
For some positive integer $d \ge 2$, events $A_1, \ldots, A_d \in \mathcal{F}$ and
a sub-$\sigma$-algebra $\mathcal{H} \subset \mathcal{F}$ are given. The events $A_i$, $i = 1, \ldots, d$,
and $\mathcal{H}$ have the following properties:
\begin{itemize}
\item[(i)] $\bigcup_{i=1}^d A_i = \Omega$.
\item[(ii)] $A_i \cap A_j = \emptyset$,\ $i, j = 1, \ldots d$, $i\neq j$.
\item[(iii)] $0 < P[A_i]$,\ $i = 1, \ldots, d$.
\item[(iv)] $0 < Q[A_i]$,\ $i = 1, \ldots, d$.
\item[(v)] $A_i \notin \mathcal{H}$,\ $i=1, \ldots, d$.
\end{itemize}
\end{Assumption}
\hl{In the literature, $P$ is also called ``source domain'' or ``training distribution'' while $Q$ is also referred to as ``target domain'' or ``test distribution'.} 

The elements $\omega$ of $\Omega$ are objects (or instances) with class (label) and covariate (or feature) attributes.
$\omega \in A_i$ means that $\omega$ belongs to class $i$ (or the positive class in the binary case if $i = 1$).

The $\sigma$-algebra $\mathcal{F}$ of events $F \in \mathcal{F}$ is a collection of subsets $F$
of $\Omega$ with the property that they can be assigned probabilities $P[F]$ and $Q[F]$
in a logically consistent way.
In the literature, thanks to their role of reflecting the available information, $\sigma$-algebras are
sometimes also called ``information set'' (Holzmann and Eulert \cite{HolzmannInformation2014}). In the following,
we use both terms exchangeably.

The sub-$\sigma$-algebra $\mathcal{H} \subset \mathcal{F}$ generated by the covariates (features) contains the events which are observable
at the time when the class of an object $\omega$ has to be predicted. Since $A_i \notin \mathcal{H}$, $i=1, \ldots, d$, then the class of the object may not yet be known.
In this paper, we assume that under the source distribution $P$, the class events $A_i$ can be observed such that the prior class probabilities can be estimated. In contrast, under the target distribution $Q$, the events $A_i$ cannot be directly observed and can only be predicted on the basis of the events $H \in \mathcal{H}$, which are assumed to reflect the features of the object.

For technical reasons, it is convenient to define the joint information set $\bal{H}$ of features and class labels:
\begin{Definition}
We denote by $\mathcal{A} = \sigma(\{A_1, \ldots, A_d\})$ the minimal sub-$\sigma$-algebra of $\mathcal{F}$ containing all $A_i$,\ $i=1, \ldots, d$ and by $\bal{H}$ the minimal sub-$\sigma$-algebra of $\mathcal{F}$ containing both $\mathcal{H}$ and $\mathcal{A}$, i.e.,\  $\bal{H} = \sigma(\mathcal{H} \cup \mathcal{A})$.
\end{Definition}
Note that the $\sigma$-algebra $\mathcal{A}$ can be represented as
\begin{subequations}
\begin{equation}\label{eq:Asigma}
\mathcal{A} \ =\
\bigl\{\bigcup_{i=1}^d (A_i \cap F_i): F_1, \ldots, F_d \in \{\emptyset, \Omega\}\bigr\},
\end{equation}
while the $\sigma$-algebra $\bal{H}$ can be written as
\begin{equation}
\bal{H}  \ =\
\bigl\{\bigcup_{i=1}^d (A_i \cap H_i): H_1, \ldots, H_d \in \mathcal{H}\bigr\}.\label{eq:Hbar}
\end{equation}
\end{subequations}

A standard assumption in machine learning is that source and target distribution are the same, i.e.,\ $P = Q$.
The situation where $P[F] \neq Q[F]$ holds for at least one $F\in \bal{H}$ is called \emph{\hl{dataset} shift} (Moreno-Torres et al.~\cite{MorenoTorres2012521}, Definition~1).

\textls[-15]{Under dataset shift as defined this way, typically, classifiers or posterior class probabilities learnt under the source distribution stop working properly under the target distribution. Finding algorithms to deal with this problem is one of the tasks in the field of \emph{{domain}~adaptation}}.

In this paper, we are mostly interested in exploring how posterior class probabilities change between a source and a target distribution as described in Assumption~\ref{as:setting}.
In particular, we provide generalisations of the posterior correction formula (2.4) of \mbox{Saerens et al.~\cite{saerens2002adjusting}} (see also Theorem~2 of Elkan \cite{Elkan01}). For this purpose, the notions of conditional expectation and conditional probability are crucial.

\hl{In the following, $E_P$ denotes conditional or unconditional expectation with respect to the probability measure $P$.} 
For a given probability space $(\Omega, \mathcal{F}, P)$, we refer to Section~8.2 of Klenke \cite{klenke2013probability} for the formal definitions and properties of
\begin{itemize}
\item The expectation
$E_P[X\,|\,\mathcal{H}]$ of a real-valued random variable $X$ conditional on a sub-$\sigma$-algebra $\mathcal{H}$;
\item The probability $P[F\,|\,\mathcal{H}]$ of an event $F \in \mathcal{F}$ conditional on $\mathcal{H}$.
\end{itemize}

In the machine learning literature, often the term \emph{posterior class probability} rather than conditional probability is used to refer to the conditional probabilities $P[A_i\,|\,\mathcal{H}]$ and $Q[A_i\,|\,\mathcal{H}]$, $i = 1, \ldots, d$, in the context of Assumption~\ref{as:setting}. In contrast, the term \emph{prior probability} is used for the probabilities $P[A_i]$ and $Q[A_i]$, which in our measure-theoretic setting should rather be called unconditional probabilities of $A_i$.

An assumption of absolute continuity is also crucial for an investigation of how the posterior class probabilities are impacted by
a change from the source distribution to the target distribution. Formally, this assumption reads as follows:
\begin{Assumption}\label{as:cont}
Assumption~\ref{as:setting} holds, and $Q$ is absolutely continuous with respect to $P$ on $\bal{H}$, i.e.,
\begin{equation*}
Q|\overline{\mathcal{H}} \ \ll\  P|\overline{\mathcal{H}},
\end{equation*}
\hl{where $M|\mathcal{H}$ stands for the measure $M$ with domain restricted to $\mathcal{H}$.} 
\end{Assumption}

The statement ``$Q$ is absolutely continuous with respect to $P$ on $\bal{H}$'' means that for all events $N \in \bal{H}$, $P[N] = 0$  implies $Q[N] = 0$. Hence, ``impossible'' events under $P$ are also impossible under $Q$. Measure-theoretic impossibility is somewhat unintuitive because for continuous distributions each single outcome has probability 0 and therefore is impossible. Nonetheless, sampled values from such distributions are single outcomes and occur despite having probability $0$.

However, the statement ``for all events $N \in \bal{H}$, $P[N] = 0$  implies $Q[N] = 0$'' is equivalent to saying: for all events $N \in \bal{H}$, $Q[N] > 0$  implies $P[N] > 0$. This means that ``possible'' events under $Q$ are also possible events under $P$, even if with
very tiny probabilities of occurrence. This phrasing of absolute continuity is more intuitive and is preferred by some authors, for instance by He et al.~\cite{he2022domain} who in \hl{Section~2} make the assumption $\mathcal{D}_T(x, y) > 0$ $\Rightarrow$  $\mathcal{D}_S(x, y) > 0$, which they seem to understand in the sense of Assumption~\ref{as:cont}.

As mentioned before, if the target distribution $Q$ is absolutely continuous with respect to $P$, there may be events whose probabilities under $Q$ are much greater than their probabilities under $P$. From a practical point of view, such events may even appear to be ``impossible'' under $P$. Notions such as ``sufficient support'' and ``support sufficiency divergence'' (\mbox{Johansson et al.~\cite{pmlr-v89-johansson19a}}) suggest that such is the view of the machine learning community.  Hence, Assumption~\ref{as:cont} is not necessarily in contrast to the working assumption of partially or fully nonoverlapping source and target domains made by many researchers in unsupervised domain adaptation.

For analyses of the case of domains where the source does not completely cover the target (such that Assumption~\ref{as:cont} may be violated), see Johannsson et al.~\cite{pmlr-v89-johansson19a}. However, the statement of Johannsson et al., \hl{Section~5}, 
``If this overlap is increased without losing information, such as through collection of additional samples, this is usually preferable.'' suggests that an assumption of nonoverlapping support is not the same as an assumption on a lack of absolute continuity.
For according to the statement by Johannsson et al., events outside of the source support do not appear to be impossible because in that case the ``collection of additional samples'' could not increase the support overlap between source and target.

Assumption~\ref{as:cont} is stronger than the common assumption of absolute continuity on $\mathcal{H}$ (see for instance, Scott \cite{Scott2019}), but in terms of interpretation there is no big difference: all events possible under the target distribution (including in label space) are also possible under the source distribution.

\textls[-5]{An important consequence of Assumption~\ref{as:cont} is that we can use the source distribution $P$ as a reference measure for the target distribution $Q$. This is more natural than introducing another measure without real-world meaning as a reference for both $P$ and $Q$. In addition, renouncing another measure as a reference has the advantageous effect of simplifying~notation.  }

Recall the following common conventions intended to make the measure-theoretic notation more incisive:
\begin{Notation}\label{not:null}
An important consequence of deploying a measure-theoretic framework as in this paper is that real-valued random variables $X$ on a fixed probability space $(\Omega, \mathcal{F}, P)$ are uniquely defined only up to events of probability $0$ and may be undefined or ill-defined on such events or when being multiplied with the factor $0$. To be more specific:
\begin{itemize}
\item If $X'$ is another random variable such that $P[X\neq X'] = 0$, then $E_P[X']$ exists if and only if $E_P[X]$ exists. In this case, $E_P[X] = E_P[X']$ follows.
\item If $X$ is undefined or ill-defined on an event $N\in\mathcal{F}$ with $P[N] = 0$, then by definition $E_p[X]$ exists if and only if $E_P[X']$ exists for
\begin{equation*}
X' = \begin{cases}
X, & \text{on}\ \Omega\setminus N,\\
0, & \text{on}\ N.
\end{cases}
\end{equation*}
In this case, $E_P[X]$ is defined as $E_P[X']$.
\item If $X$ is undefined or ill-defined on an event $F \in \mathcal{F}$ but is multiplied with another random variable $Z$ which takes the value $0$ on $F$, then, by definition, $E_p[X\,Z]$ exists if and only if $E_P[X']$ exists for
\begin{equation*}
X' = \begin{cases}
X\,Z, & \text{on}\ \Omega\setminus F,\\
0, & \text{on}\ F.
\end{cases}
\end{equation*}
In this case, $E_P[X\,Z]$ is defined as $E_P[X']$.
\end{itemize}
\end{Notation}
The conventions listed in Notation~\ref{not:null} are convenient and used frequently in the following text. Note, however, that they are only valid in the context of a fixed probability measure $P$. For instance, under Assumption~\ref{as:cont}, if the event $N$ where the random variable $X$ has probability $0$ under the source distribution $P$
of being undefined, i.e.,\ $P[N] = 0$, then $Q[N] = 0$ follows as well, such that $E_Q[X]$ should be well-defined.
Nonetheless, $Q[N]=0$ does not necessarily imply $P[N]=0$ such that $E_Q[X]$ might be well-defined despite $E_P[X]$ being ill-defined.

In the same vein, under Assumption~\ref{as:cont}, for the posterior class probabilities $P[A_i\,|\,\mathcal{H}]$, $i=1, \ldots, d$, the expectations $E_Q\bigl[P[A_i\,|\,\mathcal{H}]\bigr]$ are well-defined. However, for the posterior class probabilities $Q[A_i\,|\,\mathcal{H}]$, $i=1, \ldots, d$, the expectations $E_P\bigl[Q[A_i\,|\,\mathcal{H}]\bigr]$ are potentially ill-defined because there could be versions of $Q[A_i\,|\,\mathcal{H}]$ which are indistinguishable under $Q$ but different with positive probability under $P$. In the following, we are careful to avoid such issues whenever the discussion involves more than one probability measure.


\section{General Dataset Shift in Multinomial Classification}
\label{se:normal}

Under Assumption~\ref{as:cont}, by the Radon--Nikodym theorem, there is an $\bal{H}$-measurable density $\overline{h} = \frac{d Q}{d P}\Big|\bal{H}$ of the target distribution $Q$ with respect to  the target distribution $P$ on the joint information set $\bal{H}$ defined by \eqref{eq:Hbar}.
This density links $Q$ to $P$ by Equation~\eqref{eq:link}:
\begin{equation}\label{eq:link}
Q[F] \ = \ E_P[\overline{h}\,\mathbf{1}_F], \qquad \text{for all}\ F \in \bal{H}.
\end{equation}

In \eqref{eq:link} and in the remainder of the paper, \hl{$\mathbf{1}_F$ denotes the indicator function of $F$, defined by $\mathbf{1}_F(\omega)=1$ if $\omega \in F$ and $\mathbf{1}_F(\omega)=0$ if $\omega \not\in F$.} 

Unfortunately, in practice $\overline{h}$ is more or less unobservable. Therefore, it is desirable to decompose it into smaller parts which may be observable or can perhaps be determined through reasonable assumptions. The key step to such a decomposition is made with the following combination of definitions and lemma.

\begin{Definition}\label{de:condDist}
Under Assumption~\ref{as:setting}, define the following class-conditional distributions, by letting for $F \in \mathcal{F}$  and
$i = 1, \ldots, d$
\begin{equation}\label{eq:classcond}
P_i[F] = P[F\,|\,A_i]= \frac{P[A_i\cap F]}{P[A_i]}\quad \text{and}\quad Q_i[F] = Q[F\,|\,A_i]
= \frac{Q[A_i\cap F]}{Q[A_i]}.
\end{equation}
\end{Definition}
In the literature, when restricted to the feature information set $\mathcal{H}$, the $P_i$ and $Q_i$ sometimes are called \emph{class-conditional feature distributions}.

\begin{Lemma}\label{le:hi}
Under Assumption~\ref{as:cont}, for $i=1, \ldots, d$, the class-conditional feature distribution $Q_i$ is absolutely continuous with respect to $P_i$ on $\mathcal{H}$.

Denote by $h_i = \frac{d Q_i}{d P_i}\big|\mathcal{H}$ a Radon--Nikodym derivative (or density) of $Q_i$ with respect to $P_i$. If there is another $\mathcal{H}$-measurable function $h_i^\ast \ge 0$ with the density property, i.e., $Q_i[H] = E_{P_i}[h_i^\ast\,\mathbf{1}_H]$ for all $H\in\mathcal{H}$, then it follows that
\begin{equation}\label{eq:well}
P\bigl[h_i \neq h_i^\ast,\ P[A_i\,|\,\mathcal{H}]>0\bigr] \ = \ 0 \ = \ P[\{h_i \neq h_i^\ast\}\cap A_i].
\end{equation}
\end{Lemma}
\begin{proof}
Fix $i$ and choose any $N \in \bal{H}$  with $P_i[N] = 0$. Then, it follows that $N \cap A_i \in \bal{H}$ and $P[N \cap A_i] = 0$.
By Assumption~\ref{as:cont}, $Q[N\cap A_i] = 0$ follows, which implies
\begin{equation*}
Q_i[N] \ = \ \frac{Q[N\cap A_i]}{Q[A_i]}\ = \ 0.
\end{equation*}

Hence, we have $Q_i|\bal{H} \ll P_i|\bal{H}$ from which $Q_i|\mathcal{H} \ll P_i|\mathcal{H}$ follows.
The uniqueness of Radon--Nikodym derivatives implies
\begin{equation*}
0 \ = \ P_i[h_i \neq h_i^\ast] \ = \ \frac{P[\{h_i \neq h_i^\ast\}\cap A_i]}{P[A_i]},
\end{equation*}
and hence the right-hand side of \eqref{eq:well}.
However, by the definition of conditional probability it also follows that
\begin{equation*}
0 \ = \  P[\{h_i \neq h_i^\ast\}\cap A_i] \ = \
E_P\bigl[\mathbf{1}_{\{h_i \neq h_i^\ast\}}\,P[A_i\,|\,\mathcal{H}]\bigr].
\end{equation*}
This implies the left-hand side of \eqref{eq:well}.
\end{proof}

With Lemma~\ref{le:hi} as preparation, we are in a position to state the following key
representation result and some corollaries
for the joint density $\overline{h}$ of features and class labels. In the remainder of this paper,
we make use of \eqref{eq:form} as a normal form for $\overline{h}$.

\begin{Theorem}\label{th:dens}
Under Assumption~\ref{as:cont}, the density $\overline{h}$ of $Q$ with respect to $P$ on $\bal{H}$
can be represented as
\begin{equation}\label{eq:form}
\overline{h} \ = \ \sum_{i=1}^d h_i\,\frac{Q[A_i]}{P[A_i]}\,\mathbf{1}_{A_i}
\end{equation}
where the $h_i$ are any densities of $Q_i$ with respect to $P_i$ on $\mathcal{H}$ as introduced in
Lemma~\ref{le:hi}, for $i = 1, \ldots, d$.
\end{Theorem}
\begin{proof} Let $F \in \bal{H}$. By \eqref{eq:Hbar}, then it holds that
\begin{equation*}
F \ = \ \bigcup_{i=1}^d (A_i \cap H_i) \quad \text{for some}\ H_1, \ldots, H_d \in \mathcal{H}.
\end{equation*}
This implies
\begin{align*}
Q[F] & = \sum_{i=1}^d Q[A_i]\, Q_i[H_i]\\
& = \sum_{i=1}^d Q[A_i]\, E_{P_i}[h_i\,\mathbf{1}_{H_i}]\\
& = \sum_{i=1}^d \frac{Q[A_i]}{P[A_i]} E_P[h_i\,\mathbf{1}_{H_i\cap A_i}]\\
& = E_P\left[\left(\sum_{i=1}^d h_i\,\frac{Q[A_i]}{P[A_i]}\,\mathbf{1}_{A_i}\right)\mathbf{1}_F\right].
\end{align*}
Equation \eqref{eq:form} follows from this by the definition of Radon--Nikodym derivatives.
\end{proof}

\begin{Corollary}\label{co:densH}
Under Assumption~\ref{as:cont}, the density $h$ of $Q$ with respect to $P$ on $\mathcal{H}$ can be written as
\begin{equation*}
h \ = \ \sum_{i=1}^d h_i\,\frac{Q[A_i]}{P[A_i]}\,P[A_i\,|\,\mathcal{H}].
\end{equation*}
\end{Corollary}
\begin{proof} The corollary follows from Theorem~\ref{th:dens} because
$h = E_P[\overline{h}\,|\,\mathcal{H}]$.
\end{proof}

\begin{Corollary}\label{co:correct}
Under Assumption~\ref{as:cont}, for $i = 1, \ldots, d$, the conditional probability (posterior class probability)
$Q[A_i\,|\,\mathcal{H}]$ can be represented as
\begin{equation}\label{eq:correct}
Q[A_i\,|\,\mathcal{H}] \ = \ \frac{h_i\,\frac{Q[A_i]}{P[A_i]}\,P[A_i\,|\,\mathcal{H}]}
{\sum_{j=1}^d h_j\,\frac{Q[A_j]}{P[A_j]}\,P[A_j\,|\,\mathcal{H}]},
\end{equation}
on the set $\{h > 0\}$, where $h$ denotes the denominator of the right-hand side of \eqref{eq:correct} (and the density of $Q$ with respect to $P$ on $\mathcal{H}$, as introduced in Corollary~\ref{co:densH}).
\end{Corollary}
Equation~\eqref{eq:correct} generalises Equation~(2.4) of Saerens et al.~\cite{saerens2002adjusting} and Theorem~2 of Elkan~\cite{Elkan01} from prior probability shift to general dataset shift. Saerens et al. commented on their Equation~(2.4) as follows:
``This well-known formula can be used to compute the corrected a posteriori probabilities, $\ldots$''. Hence, in this paper we call \eqref{eq:correct} the \emph{posterior correction formula}.

Recall that under Assumption~\ref{as:cont}, it holds that $Q[h>0] = 1$ while $P[h > 0] < 1$ is possible. Hence, $Q[A_i\,|\,\mathcal{H}]$ is fully specified by \eqref{eq:correct} under $Q$ but possibly only incompletely specified under $P$.

\begin{proof}[Proof of Corollary~\ref{co:correct}]
Apply the generalised Bayes formula (see Lemma~\ref{le:GenBayes} in Appendix~\ref{se:appendix}) with
$\mathcal{F} = \bal{H}$, $f = \overline{h}$,
$\mathcal{G} = \mathcal{H}$ and $X = \mathbf{1}_{A_i}$.
\end{proof}

A direct application of the posterior correction formula \eqref{eq:correct} is not possible because the target prior probabilities $Q[A_i]$ and the target class conditional feature densities $h_i$ typically are unknown. However, in some cases the target priors might be known from external sources such as central banks, IMF or national offices of statistics. Under more specific assumptions on the type of dataset shift, it may be possible to estimate the target priors from the target dataset. See Gonz\'{a}lez et al.~\cite{Gonzalez:2017:RQL:3145473.3117807} for a survey of estimation methods under the assumption of prior probability shift.

Under prior probability shift, $h_i = 1$ is assumed for all $i$ (see Section~\ref{se:prior} below).
This means there is no change of the conditional feature distributions. This assumption might be too strong in some situations.
It might be more promising to assume similar changes for all classes (i.e.,\ $h_i \approx h_j$ for $i \neq j$), for instance, by assuming factorizable joint shift (see Section~\ref{se:factor} below), or by trying to find transformations (or representations) of the features that make the resulting feature densities similar (see Sections~\ref{se:domain} and \ref{se:GLS} below).

For the sake of completeness, we also mention the following alternative representation~\eqref{eq:alternative} of $\overline{h} = \frac{d Q}{d P}\big|\bal{H}$.
Compared to \eqref{eq:alternative}, \eqref{eq:form} provides more structural information, in particular when taking into account
Corollary~\ref{co:correct} above and, therefore, is potentially more~useful.

\begin{subequations}
\begin{Corollary}\label{co:alter}
Under Assumption~\ref{as:cont}, let $h$ be a density of $Q$ with respect to $P$ on $\mathcal{H}$.
Then, the target posterior class probabilities $Q[A_i\,|\,\mathcal{H}]$ vanish on the event $\{h > 0\}$
if the source posterior class probabilities $P[A_i\,|\,\mathcal{H}]$ vanish on $\{h > 0\}$, i.e.,\ it holds on $\{h > 0\}$ that
\begin{equation}\label{eq:vanish}
P[A_i\,|\,\mathcal{H}] = 0 \quad \Rightarrow\quad Q[A_i\,|\,\mathcal{H}] = 0.
\end{equation}
Moreover, the density $\overline{h}$ of $Q$ with respect to $P$ on $\bal{H}$ can be represented as
\begin{equation}\label{eq:alternative}
\overline{h} \ = \ h \sum_{i=1}^d \frac{Q[A_i\,|\,\mathcal{H}]}{P[A_i\,|\,\mathcal{H}]}
\mathbf{1}_{A_i}.
\end{equation}
\end{Corollary}
\end{subequations}

\begin{proof}
Equation~\eqref{eq:vanish} follows immediately from Corollary~\ref{co:correct}.
Taking into account Notation~\ref{not:null} for the meaning of \eqref{eq:alternative} on the event $\{P[A_i\,|\,\mathcal{H}] = 0\}$, the equation follows from \eqref{eq:Hbar} and the definition of the posterior class probabilities.
\end{proof}

The following result may be considered an inversion of the previous results and in particular Corollary~\ref{co:correct} on
the relationship between source and target distributions. It is of interest mostly for dealing with sample selection bias (see Section~\ref{se:sample} below).

\begin{Proposition}\label{pr:reverse}
In the setting of Theorem~\ref{th:dens}, assume additionally that $P[\overline{h}=0] = 0$ holds. Then, the following statements hold true:
\begin{itemize}
\item [(i)] $P$ is absolutely continuous with respect to $Q$ on $\bal{H}$, with $\frac{d P}{d Q}\big|\bal{H} = 1 / \overline{h}$.
\item[(ii)] For $i=1,\ldots, d$, the source class-conditional feature distribution $P_i$ is absolutely continuous with respect to $Q_i$ on $\mathcal{H}$, with $Q_i[h_i = 0] =0 = P[h_i=0]$ and
\begin{equation*}
\frac{d P_i}{d Q_i}\Big|\mathcal{H} \ = \ \frac{1}{h_i}.
\end{equation*}
\item[(iii)] The density $\frac{d P}{d Q}\big|\bal{H}$ can also be represented as
\begin{equation*}
\frac{d P}{d Q}\Big|\bal{H}  \ = \
\sum_{i=1}^d \frac{1}{h_i}\,\frac{P[A_i]}{Q[A_i]}\,\mathbf{1}_{A_i}.
\end{equation*}
\item [(iv)] The density $\frac{d P}{d Q}\big|\mathcal{H}$ can be represented as
\begin{equation*}
\frac{d P}{d Q}\Big|\mathcal{H} \ = \
\sum_{i=1}^d
\frac{1}{h_i}\,\frac{P[A_i]}{Q[A_i]}\,Q[A_i\,|\,\mathcal{H}].
\end{equation*}
\item [(v)] For $i=1,\ldots, d$, it holds that
\begin{equation*}
P[A_i\,|\,\mathcal{H}] \ = \ \frac{\frac{1}{h_i}\,Q[A_i\,|\,\mathcal{H}]\,\frac{P[A_i]}{Q[A_i]}}
{\sum_{j=1}^d \frac{1}{h_j}\,Q[A_j\,|\,\mathcal{H}]\,\frac{P[A_j]}{Q[A_j]}}.
\end{equation*}
\end{itemize}
\end{Proposition}

\begin{proof}
(i) is a well-known property of equivalent probability measures (see Problem 32.6 of Billingsley \cite{billingsley1986probability}).

By (i), $P$ is absolutely continuous with respect to $Q$ on $\bal{H}$. This implies that $P_i$ is absolutely continuous with respect to $Q_i$ on $\mathcal{H}$ and, again by Problem 32.6 of \cite{billingsley1986probability}, the rest of (ii) follows as well.

Properties (iii), (iv) and (v) follow from (i) and (ii), by making use of Theorem~\ref{th:dens} and Corollaries~\ref{co:densH} and \ref{co:correct} with swapped roles of $P$ and $Q$.
\end{proof}


\section{Factorizable Joint Shift}
\label{se:factor}

The following definition translates Definition~2.2 of He et al.~\cite{he2022domain} into the setting of this~paper.
\begin{subequations}
\begin{Definition}\label{de:FJS}
Under Assumption~\ref{as:cont}, we say that the target distribution $Q$ is related to the source distribution $P$ by \emph{factorizable joint shift (FJS)}, if there are a non-negative $\mathcal{H}$-measurable function $g$ and a non-negative $\mathcal{A}$-measurable function $b$ such that the density $\overline{h}$ of $Q$ with respect to $P$ on $\bal{H}$ can be represented as
\begin{equation}\label{eq:factorized}
\overline{h} \ = \ g\,b.
\end{equation}
\end{Definition}

Observe that the functions $g$ and $b$ of Definition~\ref{de:FJS} are not uniquely determined because for any $c>0$ the functions $g_c = c\,g$ and $b_c = b / c$ are also $\mathcal{H}$-measurable and $\mathcal{A}$-measurable, respectively, and satisfy
\begin{equation}\label{eq:nonunique}
\overline{h} \ = \ g_c\,b_c.
\end{equation}
\end{subequations}
In the remainder of this section, we show that the functions $g$ and $b$ depend on the source distribution $P$ as well as the marginal distributions of $Q$ on $\mathcal{H}$ and $\mathcal{A}$, respectively, but not on the joint distribution $Q|\bal{H}$.
For the case $d=2$, in Section~\ref{se:binary} below we obtain the stronger result that
\begin{itemize}
\item $g$ and $b$ are uniquely determined (up to the ambiguity expressed by \eqref{eq:nonunique}) by the marginal distributions of $Q$ on $\mathcal{H}$ and $\mathcal{A}$ and the source distribution $P$;
\item With fixed source distribution $P$, for each pair of marginal distributions of $Q$ on $\mathcal{H}$ and $\mathcal{A}$, there exists (up to a constant factor) a factorization \eqref{eq:factorized}.
\end{itemize}

\begin{Theorem}\label{th:factorized}
Under Assumption~\ref{as:cont}, let the source distribution $P$ and the target distribution $Q$ be related by joint factorizable shift in the sense of Definition~\ref{de:FJS}.
Denote by $h$ the density of $Q$ with respect to $P$ on $\mathcal{H}$ and let $q_i = Q[A_i]$ and $p_i = P[A_i]$, $i = 1, \ldots, d$.

Then, up to a constant factor $c$ as in \eqref{eq:nonunique}, it follows that
\begin{subequations}
\begin{align}\label{eq:b}
b & = \sum_{i=1}^{d-1} \varrho_i\,\frac{q_i}{p_i}\,\mathbf{1}_{A_i} +
\frac{q_d}{p_d}\,\mathbf{1}_{A_d}\quad \text{and}\\
g & = \frac{h}{\sum_{i=1}^{d-1} \varrho_i\,\frac{q_i}{p_i}\,P[A_i\,|\,\mathcal{H}] +
\frac{q_d}{p_d}\,P[A_d\,|\,\mathcal{H}]},\label{eq:g}
\end{align}
where the constants $\varrho_1, \ldots, \varrho_{d-1}$ are positive and finite and satisfy the following equation system:
\begin{equation}\label{eq:system}
p_j \ = \ \varrho_j\,E_P\left[\frac{h\,P[A_j\,|\,\mathcal{H}]}
{\sum_{i=1}^{d-1} \varrho_i\,\frac{q_i}{p_i}\,P[A_i\,|\,\mathcal{H}] +
\frac{q_d}{p_d}\,P[A_d\,|\,\mathcal{H}]}\right], \quad j = 1, \ldots, d-1.
\end{equation}
\end{subequations}

Conversely, let an $\mathcal{H}$-measurable function $h \ge 0$ with $E_P[h] =1$ and $(q_i)_{i=1, \ldots, d} \in (0,1)^d$ with $\sum_{i=1}^d q_i =1$ be given. If $\varrho_1 >0$, $\ldots$, $\varrho_{d-1} > 0$ are solutions of the equation system \eqref{eq:system} and $b$ and $g$ are defined by \eqref{eq:b} and \eqref{eq:g}, respectively, then $g\,b$ is a density of a probability measure $Q$ with respect to $P$ on $\bal{H}$, such that $h$ is the marginal density of $Q$ with respect to $P$ on $\mathcal{H}$
and $Q[A_i] = q_i$ holds for $i = 1, \ldots, d$.
\end{Theorem}
\begin{proof}
First, we show that \eqref{eq:b}--\eqref{eq:system} are necessary if $Q$ and $P$ are related by factorizable joint shift as in \eqref{eq:factorized}.

Since $b$ is $\mathcal{A}$-measurable by assumption, there are constants $\beta_1, \ldots, \beta_d\in\mathbb{R}$ such that
\begin{subequations}
\begin{equation}\label{eq:bfirst}
b \ = \ \sum_{i=1}^d \beta_i\,\mathbf{1}_{A_i}.
\end{equation}

For fixed $k\in\{1, \ldots, d\}$, this implies
\begin{equation*}
q_k  = E_P[g\,b\,\mathbf{1}_{A_k}]
= \beta_k\,E_P[g\,\mathbf{1}_{A_k}]
= \beta_k\,p_k\,E_{P_k}[g].
\end{equation*}

By Assumption~\ref{as:cont}, we have $q_k > 0$ and $p_k>0$. Hence, it follows $E_{P_k}[g]>0$ and
\begin{equation}\label{eq:beta}
\beta_k \ = \ \frac{q_k}{p_k\,E_{P_k}[g]}\ > \ 0.
\end{equation}
\end{subequations}

As $g\,b$ is by assumption an $\bal{H}$-density of $Q$ with respect to $P$, it follows that
\begin{equation*}
h \ = \ E_p[g\,b\,|\,\mathcal{H}] \ = \ g\,\sum_{i=1}^d \frac{q_i}{p_i\,E_{P_i}[g]}\,P[A_i\,|\,\mathcal{H}].
\end{equation*}

The relations $1 = \sum_{i=1}^d P[A_i\,|\,\mathcal{H}]$ and \eqref{eq:beta} imply
\begin{equation*}
\sum_{i=1}^d \frac{q_i}{p_i\,E_{P_i}[g]}\,P[A_i\,|\,\mathcal{H}] \ > \ 0.
\end{equation*}

Therefore, we obtain
\begin{equation}\label{eq:interim}
g \ = \ \frac{h}{\sum_{i=1}^d \frac{q_i}{p_i\,E_{P_i}[g]}\,P[A_i\,|\,\mathcal{H}]}.
\end{equation}

For $k \in \{1, \ldots, d-1\}$, \eqref{eq:interim} implies
\begin{align*}
E_{P_k}[g] & = \frac{E_P\bigl[g\,P[A_k\,|\,\mathcal{H}]\bigr]}{p_k}\\
& = \frac{1}{p_k} E_P\left[\frac{h\,P[A_k\,|\,\mathcal{H}]}
{\sum_{i=1}^d \frac{q_i}{p_i\,E_{P_i}[g]}\,P[A_i\,|\,\mathcal{H}]} \right],
\end{align*}
and, equivalently,
\begin{equation*}
p_k  = \frac{E_{P_d}[g]}{E_{P_k}[g]}
E_P\left[\frac{h\,P[A_k\,|\,\mathcal{H}]}
{\sum_{i=1}^{d-1} \frac{E_{P_d}[g]}{E_{P_i}[g]}\, \frac{q_i}{p_i}\,P[A_i\,|\,\mathcal{H}] +
\frac{q_d}{p_d}\,P[A_d\,|\,\mathcal{H}]} \right].
\end{equation*}

With $\varrho_k = \frac{E_{P_d}[g]}{E_{P_k}[g]} > 0$, this implies \eqref{eq:system}.
Equations \eqref{eq:b} and \eqref{eq:g} follow from multiplying~\eqref{eq:bfirst} with $E_{P_d}[g]$ and \eqref{eq:interim} with $1/E_{P_d}[g]$, respectively.

The converse statement follows from the following observations:
\begin{itemize}
\item With $b$ and $g$ as in \eqref{eq:b} and \eqref{eq:g}, $E_P[g\,b] = 1$ holds such that $g\,b$ is an $\bal{H}$-measurable density with respect to $P$.
\item Furthermore, $E_P[g\,b\,|\,\mathcal{H}] = h$ holds such that $h$ is the marginal density of $g\,b$ on $\mathcal{H}$ with respect to $P$.
\item For $j \in \{1, \ldots, d-1\}$, \eqref{eq:system} is actually equivalent to
\begin{equation*}
Q[A_j] \ = \ E_P[g\,b\,\mathbf{1}_{A_j}] \ = \ q_j.
\end{equation*}
\end{itemize}

Finally, $q_d = Q[A_d]$ is implied by $\sum_{i=1}^d q_i = 1$.
\end{proof}

Thanks to Theorem~\ref{th:factorized}, the following version of the posterior correction formula \eqref{eq:correct} can be given for factorizable joint shift.
\begin{Corollary}\label{co:factorized}
Under Assumption~\ref{as:cont}, let the source distribution $P$ and the target distribution $Q$ be related by joint factorizable shift in the sense of Definition~\ref{de:FJS}.
Denote by $h$ the density of $Q$ with respect to $P$ on $\mathcal{H}$. Then, the target posterior probabilities $Q[A_j\,|\,\mathcal{H}]$, $j = 1, \ldots, d$, can be represented as functions of the source posterior probabilities $P[A_j\,|\,\mathcal{H}]$, $j = 1, \ldots, d$, in
the following way on the event $\{h > 0\}$:
\begin{equation}\label{eq:FJScorrect}
\begin{split}
Q[A_j\,|\,\mathcal{H}] & = \frac{\varrho_j\,\frac{Q[A_j]}{P[A_j]} P[A_j\,|\,\mathcal{H}]}
{\sum_{i=1}^{d-1} \varrho_i\,\frac{Q[A_i]}{P[A_i]} P[A_i\,|\,\mathcal{H}] +
\frac{Q[A_d]}{P[A_d]} P[A_d\,|\,\mathcal{H}]},\quad j = 1, \ldots, d-1,\\
Q[A_d\,|\,\mathcal{H}] & = \frac{\frac{Q[A_d]}{P[A_d]} P[A_d\,|\,\mathcal{H}]}
{\sum_{i=1}^{d-1} \varrho_i\,\frac{Q[A_i]}{P[A_i]} P[A_i\,|\,\mathcal{H}] +
\frac{Q[A_d]}{P[A_d]} P[A_d\,|\,\mathcal{H}]},
\end{split}
\end{equation}
where the positive constants $\varrho_1, \ldots, \varrho_{d-1}$ satisfy the equation system \eqref{eq:system}.
\end{Corollary}
\begin{proof}
Apply the generalised Bayes formula (Lemma~\ref{le:GenBayes} in Appendix~\ref{se:appendix}) for $\mathcal{G} = \mathcal{H}$,
$X = \mathbf{1}_{A_j}$ and $f = g\,b$, with $g$ and $b$ specified by \eqref{eq:g} and \eqref{eq:b}, respectively.
\end{proof}
\begin{Remark}\label{rm:invariant}
Assuming $P[A_d\,|\,\mathcal{H}] > 0$, \eqref{eq:FJScorrect} implies
\begin{equation}\label{eq:ratio}
\frac{Q[A_j\,|\,\mathcal{H}]}{Q[A_d\,|\,\mathcal{H}]} \frac{Q[A_d]}{Q[A_j]} \ = \
\varrho_j\, \frac{P[A_j\,|\,\mathcal{H}]}{P[A_d\,|\,\mathcal{H}]} \frac{P[A_d]}{P[A_j]},
\quad j = 1, \dots, d-1.
\end{equation}
Recall that $P[A_k\,|\,\mathcal{H}] / P[A_k]$ is the density with respect to $P$ of the class-conditional feature distribution $P_k$, as defined by \eqref{eq:classcond}, on the feature information set $\mathcal{H}$. Similarly, $Q[A_k\,|\,\mathcal{H}] / Q[A_k]$ is the density with respect to $Q$ of the class-conditional feature distribution $Q_k$ on $\mathcal{H}$. Therefore, \eqref{eq:ratio} states that
under factorizable joint shift, the ratios of the class-conditional feature densities are invariant up to a constant factor.
\end{Remark}
Remark~\ref{rm:invariant} suggests joint factorizable shift could also be called \emph{scaled density ratios} shift. This term would emphasise a probabilistic interpretation of this kind of dataset shift, in contrast to ``factorizable joint shift'' with its focus
on the technical aspect of separation of input and output variables.

\subsection{Alternatives to Joint Importance Aligning}
\label{se:alternative}

He et al.~\cite{he2022domain} proposed in \hl{Section~3} the ``joint importance aligning'' method for estimating a factorized version of the ratio of source and target domain densities which they called ``joint importance weight''. He et al.\ presented a ``supervised'' and an ``unsupervised'' version of their method. The ``unsupervised'' version was intended for the case where no class labels were observed
in the target domain, i.e.,\ the case considered primarily in this~paper. 

Regarding the performance of the ``unsupervised'' version of their proposal, He et al.\ indicated that the proposed method tended to
present simple covariate shift (see Section~\ref{se:covshift} below) as a solution. This does not come as a surprise because He et al.~\cite{he2022domain} stated ``$\ldots$ in unsupervised objective, we define $\tilde{V}(x) \triangleq \mathbb{E}_{y\sim \mathcal{D}_S(y|x)} V(y)$  $\dots$'', which suggests that the authors implicitly assumed $\mathcal{D}_S(y|x) = \mathcal{D}_T(y|x)$, i.e.,\ covariate shift.
Without providing an explanation, He et al.\ proposed a discretisation of the data (covariate) space in order to prevent the algorithm from converging to covariate shift as solution.

Given these qualms about ``joint importance aligning', it might be useful to point out alternative approaches to finding the factorization \eqref{eq:factorized}, based on Theorem~\ref{th:factorized}.
The theorem suggests two obvious ways to learn the characteristics of factorizable joint shift:
\begin{itemize}
\item[(a)] If the target prior class probabilities $Q[A_i]$ are known (for instance from external sources), solve \eqref{eq:system} for the constants $\varrho_i$.
\item[(b)] If the target prior class probabilities $Q[A_i]$ are unknown, fix values for the constants $\varrho_i$ and solve \eqref{eq:system} for the $Q[A_i]$. Letting $\varrho_i =1$ for all $i$ is a natural choice that converts~\eqref{eq:system} into the system of maximum likelihood equations for the $Q[A_i]$ under the prior probability shift assumption.
\end{itemize}

See Section~4.2.4 of Tasche \cite{tasche2013art} for an example of approach~(a) from the area of credit risk.
Regarding the interpretation of \eqref{eq:system} in approach~(b) as maximum likelihood equations, see Du Plessis and Sugiyama \cite{duPlessis2014110} or Tasche \cite{tasche2013law}.
This interpretation, in particular, implies that an EM (expectation maximisation) algorithm can be deployed for solving the equation
system (Saerens et al.\ \cite{saerens2002adjusting}).

\subsection{The Binary Case}
\label{se:binary}

Theorem~\ref{th:factorized} does not provide sufficient or necessary conditions for the existence or uniqueness of solutions to equation system \eqref{eq:system} if a density $h$ and a candidate class distribution $(q_i)_{i=1,\ldots,d}$ are given. In the special case $d=2$, such an existence and uniqueness statement can be made as the following proposition shows.
The following proposition is a generalisation of Section~4.2.4 of Tasche \cite{tasche2013art}.
\begin{Proposition}\label{pr:d2}
Let $(\Omega, \mathcal{F}, P)$ be a probability space, $\mathcal{H} \subset \mathcal{F}$ a sub-$\sigma$-algebra of $\mathcal{F}$ and $A \in \mathcal{F}\setminus\mathcal{H}$ with $0 < p = P[A] < 1$. Assume that $P\bigl[P[A\,|\,\mathcal{H}]\in\{0,1\}\bigr] = 0$.

Then, there exists a solution $\varrho = \varrho_1 > 0$ to \eqref{eq:system} with $A_1 = A$, $A_2 = \Omega\setminus A$, $p_1 = p = 1-p_2$ and $q_1 = q = 1-q_2$, if an $\mathcal{H}$-measurable function $h: \Omega \to [0,\infty)$ with $E_P[h] = 1$ and a number $0 < q < 1$ are given.

Assume additionally that $\mathcal{H}$ and $A$ are not independent under $P$.
Then, the solution $\varrho$ to \eqref{eq:system} is unique. Denote by $\phi:(0,1)\to (0, \infty)$ the function that maps, for a fixed density $h$, the number $0 < q < 1$ to $\varrho$, i.e.,\ $\phi(q)=\varrho$.
Then, $\phi$ has the following properties:
\begin{itemize}
\item[(i)] $\phi$ is strictly increasing and continuous on $(0,1)$.
\item[(ii)] $\lim\limits_{q\to 0} \phi(q) = \frac{P[A]}{(1-P[A])\,
E_P\left[h\,\frac{P[A\,|\,\mathcal{H}]}{1-P[A\,|\,\mathcal{H}]}\right]}$.
\item[(iii)] $\lim\limits_{q\to 1} \phi(q)  = \frac{P[A]}{1-P[A]}
E_P\left[h\,\frac{1-P[A\,|\,\mathcal{H}]}{P[A\,|\,\mathcal{H}]}\right]$.
\end{itemize}
\end{Proposition}
See Appendix~\ref{se:proofs} for a proof of Proposition~\ref{pr:d2}.
The uniqueness statement of Proposition~\ref{pr:d2} is interesting because it implies an answer to the question of whether proper \emph{concept shift} (dataset shift where the marginal distributions of the features and labels, respectively, remain unchanged) can be modelled as factorizable joint shift. The answer---at least for the binary case---is no, because ``no shift'' then provides the only solution to Equation~\eqref{eq:system}.

\section{Common Types of Dataset Shift}
\label{se:examples}

In this section, we revisit some popular special cases of dataset shift. In each case, we discuss the question if factorizable joint shift is implied or if the special type of shift is implied by factorizable joint shift. In addition, we provide in each case an adapted version of the posterior correction formula \eqref{eq:correct}.

\subsection{Prior Probability Shift}
\label{se:prior}

\textls[-15]{Moreno-Torres et al.~\cite{MorenoTorres2012521} defined \emph{prior probability shift} as invariance of the class-conditional} feature distributions between source and target, i.e.,
\begin{subequations}
\begin{equation}\label{eq:priorShift}
Q_i[H] \ = \ P_i[H], \qquad H \in \mathcal{H}, i = 1, \ldots, d,
\end{equation}
with $Q_i$ and $P_i$ defined as in \eqref{eq:classcond} above, and $Q[A_i] \neq P[A_i]$ for at least one $i$.
\hl{%
This type of dataset shift is also known as ``target shift''} \cite{Zhang:2013:TargetShift}, ``global drift'' \cite{hofer2013drift}, ``label shift'' \cite{pmlr-v80-lipton18a} and under other names. 
In terms of the notation used in Theorem~\ref{th:dens}, \eqref{eq:priorShift} is equivalent to having the densities of the $Q_i$ with respect to the $P_i$ on the feature
information set $\mathcal{H}$ equal to $1$, i.e.,
\begin{equation}\label{eq:priorhi}
h_i \ = \ 1, \qquad i = 1, \ldots, d.
\end{equation}
\end{subequations}

By Theorem~\ref{th:dens}, \eqref{eq:priorhi} implies for the density $\overline{h}$ of $Q$ with respect to $P$ on $\bal{H}$ that
\begin{equation}\label{eq:priordens}
\overline{h} \ = \ \sum_{i=1}^d \frac{Q[A_i]}{P[A_i]}\,\mathbf{1}_{A_i}
\ = \ \frac{\sum_{i=1}^d Q[A_i]\,\mathbf{1}_{A_i}}{\sum_{i=1}^d P[A_i]\,\mathbf{1}_{A_i}},
\end{equation}
which obviously is an $\mathcal{A}$-measurable function. Definition~\ref{de:FJS} of factorizable joint shift, therefore, is satisfied---as stated by He et al.~\cite{he2022domain} in \hl{Table~1}.

The posterior correction formula \eqref{eq:correct} in this case takes the well-known form
\begin{equation}\label{eq:priorcorrect}
Q[A_i\,|\,\mathcal{H}] \ = \ \frac{\frac{Q[A_i]}{P[A_i]}\,P[A_i\,|\,\mathcal{H}]}
{\sum_{j=1}^d \frac{Q[A_j]}{P[A_j]}\,P[A_j\,|\,\mathcal{H}]},
\end{equation}
as noted before, e.g., by Saerens et al.~\cite{saerens2002adjusting} and Elkan \cite{Elkan01}.

\subsection{Covariate Shift}
\label{se:covshift}

Moreno-Torres et al.~\cite{MorenoTorres2012521} defined \emph{covariate shift} as invariance of the posterior class probabilities between source and target, i.e.,
\begin{equation}\label{eq:covariateShift}
Q[A_i\,|\,\mathcal{H}] \ = \ P[A_i\,|\,\mathcal{H}], \qquad  i = 1, \ldots, d,
\end{equation}
and $Q[H] \neq P[H]$ for at least one $H \in \mathcal{H}$.

\begin{Proposition}\label{pr:equiv}
Under Assumption~\ref{as:cont}, denote by $\overline{h}$ and $h$, as in Section~\ref{se:normal}, the densities of $Q$ with respect to $P$ on $\bal{H}$ and $\mathcal{H}$, respectively.
Then, \eqref{eq:covariateShift} holds true if and only if $h$ is also a density of $Q$ with respect to $P$ on $\bal{H}$, i.e.,\ $P[\overline{h}=h] = 1$.
\end{Proposition}
\begin{proof}
The ``if'' part of the assertion is Lemma~1 of Tasche \cite{tasche2022class}. Taking into account Notation~\ref{not:null}, the ``only if'' is implied by Corollary~\ref{co:alter}.
\end{proof}

Proposition~\ref{pr:equiv} implies that covariate shift is a special case of factorizable joint shift in the sense of Definition~\ref{de:FJS}, with $b=1$ and $g=h$, as noted in Table~1 of He et al.~\cite{he2022domain}.

Then, observe that the fact that $b$ is constant implies by \eqref{eq:b} that
\begin{equation}\label{eq:rhoi}
\varrho_i\ = \ \frac{Q[A_d]}{P[A_d]}\,\frac{P[A_i]}{Q[A_i]}, \qquad \text{for all}\ i=1, \ldots, d-1.
\end{equation}

It can readily be checked that under the assumption of covariate shift the $\varrho_i$ defined by \eqref{eq:rhoi} indeed solve equation system \eqref{eq:system}.

\subsection{Covariate Shift with Posterior Drift}

Scott \cite{Scott2019} defined \emph{covariate shift with posterior drift} (CSPD) for the binary special case ($d=2$) of Assumption~\ref{as:setting} as the following variant of \eqref{eq:covariateShift}:

there exists a strictly increasing function $\varphi$ such that
\begin{equation}\label{eq:CSPD}
Q[A_1\,|\,\mathcal{H}] \ = \ \varphi\bigl(P[A_1\,|\,\mathcal{H}]\bigr).
\end{equation}

Equation \eqref{eq:CSPD} implies that $Q[A_1\,|\,\mathcal{H}]$ and $P[A_1\,|\,\mathcal{H}]$ are strongly comonotonic. As shown in Tasche \cite{Tasche2022}, the converse implication also holds true.

Note that from \eqref{eq:CSPD}, it also follows that
\begin{equation*}
Q[A_2\,|\,\mathcal{H}] \ = \ 1-\varphi\bigl(1-P[A_2\,|\,\mathcal{H}]\bigr).
\end{equation*}

Hence, the increasing link between the posterior positive class probabilities defining CSPD does not only apply to class $A_1$ but automatically also to the negative class $A_2$.

CSPD is implied by factorizable joint shift. This follows from \eqref{eq:FJScorrect} because of
\begin{equation*}
Q[A_2\,|\,\mathcal{H}] \ = \ \varphi^\ast\bigl(P[A_2\,|\,\mathcal{H}]\bigr),
\end{equation*}
with $\varphi^\ast(x) = \frac{\frac{Q[A_2]}{P[A_2]}\,x}{\varrho_1\,\frac{Q[A_1]}{P[A_1]}\,(1-x) +
\frac{Q[A_2]}{P[A_2]}\,x}$ which is strictly increasing in $x$.

Under CSPD, the class-conditional densities $h_i = \frac{d Q_i}{d P_i}\big|\mathcal{H}$, $i=1,2$, introduced in Lemma~\ref{le:hi} can be shown to be
\begin{equation}\label{eq:hiCSPS}
\begin{split}
h_1 & = \frac{Q[A_1]}{P[A_1]}\,h\,\frac{\varphi\bigl(P[A_1\,|\,\mathcal{H}]\bigr)}{P[A_1\,|\,\mathcal{H}]\bigr)}\quad
\text{and}\\
h_2 & = \frac{1-Q[A_1]}{1-P[A_1]}\,h\,
\frac{1-\varphi\bigl(P[A_1\,|\,\mathcal{H}]\bigr)}{1-P[A_1\,|\,\mathcal{H}]\bigr)},
\end{split}
\end{equation}
where $h$ is the density of $Q$ with respect to $P$ on the feature information set $\mathcal{H}$.
Alas, when used in connection with Theorem~\ref{th:dens}, \eqref{eq:hiCSPS} does not provide a very useful representation of $\overline{h}$.

\subsection{Domain Invariance}
\label{se:domain}

Translated into the concepts and notation of this paper, \emph{domain invariance} (see Table~1 of He et al.~\cite{he2022domain}) is defined as follows:
\begin{subequations}
\begin{itemize}
\item There is an $\mathcal{H}$-measurable mapping (transformation) $T$ into some measurable space with the property that
\begin{equation}\label{eq:GPQ}
Q[M] \ = \ P[M]
\qquad \text{for all}\ M \in \sigma\bigl(\mathcal{A}\cup \mathcal{G}\bigr),
\end{equation}
\textls[-20]{where $\mathcal{G}=\sigma(T)$ denotes the smallest sub-$\sigma$-algebra of $\mathcal{H}$ such that $T$ is still \mbox{$\mathcal{G}$-measurable.}}
\item For all $i = 1, \ldots d$ it holds that:
\begin{equation}\label{eq:sufficiency}
Q[A_i\,|\,\mathcal{H}] \ = \ P[A_i\,|\,\mathcal{G}]\quad
\text{and}\quad Q[A_i\,|\,\mathcal{H}] \ = \ Q[A_i\,|\,\mathcal{G}].
\end{equation}
\end{itemize}

Property \eqref{eq:sufficiency} means that $T$ is \emph{sufficient} for $\mathcal{H}$ under both $P$ and $Q$ in the sense of Section~32.3 of Devroye et al.~\cite{devroye1996probabilistic}.

As mentioned in He et al.~\cite{he2022domain}, \eqref{eq:GPQ} implies covariate shift with respect to $\mathcal{G}$, i.e.,
\begin{equation}\label{eq:covG}
Q[A_i\,|\,\mathcal{G}] \ = \ P[A_i\,|\,\mathcal{G}], \quad i = 1, \ldots, d.
\end{equation}
\end{subequations}

From \eqref{eq:sufficiency} then follows covariate shift with respect to $\mathcal{H}$.

Actually, this reasoning shows that in the definition of domain invariance according to He et al.~\cite{he2022domain}, \eqref{eq:GPQ} could be replaced by the weaker assumption \eqref{eq:covG}, without losing the consequence that covariate shift holds on the whole information set $\mathcal{H}$.

\subsection{Generalised Label Shift}
\label{se:GLS}

\textls[-15]{Tachet des Combes et al.~\cite{tachetdescombes2020domain} defined \emph{generalised label shift (GLS)} as follows: there is an $\mathcal{H}$-measurable mapping (transformation) $T$ into some measurable space  with the property~that}
\begin{equation}\label{eq:GLS}
Q[G\,|\,A_i] \ =\ P[G\,|\, A_i], \qquad i = 1, \ldots, d, G \in \mathcal{G}=\sigma(T).
\end{equation}

Since $\sigma(T) \subset \mathcal{H}$ holds, this is weaker than requiring~\eqref{eq:priorShift} as for prior probability shift. In this sense, GLS generalises prior probability shift.

He et al.~\cite{he2022domain} gave in Table~1 a narrower definition of GLS, by requiring in addition to~\eqref{eq:GLS} also \eqref{eq:sufficiency}, and went on to prove that GLS implied factorizable joint shift.
We provide an alternative proof of this result, providing mathematically rigorous meaning for the factorisation proposed by He et al.

\begin{Proposition}\label{pr:GLS}
Under Assumption~\ref{as:cont}, let there be an $\mathcal{H}$-measurable mapping $T$ into some measurable space such that \eqref{eq:GLS} and \eqref{eq:sufficiency} hold. Denote by $h$ the density of the target distribution $Q$ with respect to the source distribution $P$ on $\mathcal{H}$. Then, $Q$ and $P$ are related by factorizable joint shift in the sense of Definition~\ref{de:FJS}, with
\begin{equation}\label{eq:GLSsplit}
\begin{split}
b & = \sum_{i=1}^d \frac{Q[A_i]}{P[A_i]} \mathbf{1}_{A_i} \quad \text{and}\\
g & = \frac{h}{\sum_{i=1}^d \frac{Q[A_i]}{P[A_i]} P[A_i\,|\,\mathcal{H}]}.
\end{split}
\end{equation}
\end{Proposition}

See Appendix~\ref{se:proofs} for a proof of Proposition~\ref{pr:GLS}. Observe that Proposition~\ref{pr:GLS} and Corollary~\ref{co:factorized} together imply that the same class posterior correction formula \eqref{eq:priorcorrect} applies for generalised label shift and prior probability shift.

The factorisation presented in \eqref{eq:GLSsplit} of Proposition~\ref{pr:GLS} corresponds to the factorisation of generalised label shift proposed by He et al.~\cite{he2022domain} in Table~1 in the following way:
\begin{itemize}
\item Function $b$ matches $\mathcal{D}_T(Y)$ of He et al. Because in this paper the reference measure is the source distribution $P$, $\mathcal{D}_S(Y)$ of He et al.\ corresponds to constant $1$ in \eqref{eq:GLSsplit}.
\item Function $h$ matches $\mathcal{D}_T(X)$  and function $\gamma$ (the denominator of $g$) matches $\mathcal{D}_T(Z)$ of He et al. Due to our reference measure being $P$, $\mathcal{D}_S(X)$ and $\mathcal{D}_S(Z)$ of He et al.\ are both matched by constant $1$. The term
$\mathcal{D}_T(X=x\,|\,Z=g(x))$ appears in \eqref{eq:GLSsplit} as the density ratio $g = h/\gamma$, hence with a well-defined mathematical meaning.
\end{itemize}

\begin{Remark}
Proposition~\ref{pr:GLS} combined with Remark~\ref{rm:invariant} shows that ``generalized label shift'' in the sense of He et al.~\cite{he2022domain} is the same type of dataset shift that was discussed as ``invariant density ratio''-type dataset shift in Tasche~\cite{tasche2017fisher}.
\end{Remark}


\section{Sample Selection Bias}
\label{se:sample}

Sample selection bias is an important cause of dataset shift. In this subsection, we revisit parts of Hein \cite{Hein2009binary} in order to illustrate some of the concepts and results presented before. We basically work under Assumption~\ref{as:setting} but without the interpretation of $P$ as source and $Q$ as target distribution. Instead, $P$ is interpreted as the distribution of a population from
which a potentially biased random sample is taken, resulting in the distribution $Q$. When studying sample selection bias in this setting, the goal is to infer properties of $P$ from properties of the sample distribution $Q$.

The following assumption describes the setting of this section. The idea is that under the population distribution, each object has a positive chance to be selected.
This chance may depend upon the features (covariates) and the class of the object.

\begin{Assumption}[Sample selection]\label{as:select}
$(\Omega, \mathcal{F})$ is a measurable space. The \emph{population distribution} $P$ is a probability measure on $(\Omega,\mathcal{F})$.
For some positive integer $d \ge 2$, events $A_1, \ldots, A_d \in \mathcal{F}$ and a sub-$\sigma$-algebra $\mathcal{H} \subset \mathcal{F}$ are given. The events $A_i$, $i = 1, \ldots, d$ and $\mathcal{H}$ have the following properties:
\begin{itemize}
\item[(i)] $\bigcup_{i=1}^d A_i = \Omega$.
\item[(ii)] $A_i \cap A_j = \emptyset$,\ $i, j = 1, \ldots d$, $i\neq j$.
\item[(iii)] $0 < P[A_i]$,\ $i = 1, \ldots, d$.
\item[(iv)] $A_i \notin \mathcal{H}$,\ $i=1, \ldots, d$.
\end{itemize}

The \emph{selection probability} is an $\bal{H}$-measurable random variable $0 < \varphi \le 1$ where the sub-$\sigma$-algebra $\bal{H}$ is defined as in \eqref{eq:Hbar}.

The probability space $(\Omega, \mathcal{F}, P)$ also supports a random variable $U$ which is uniformly distributed on $[0,1]$ 
such that $U$ and $\bal{H}$ are independent.
\end{Assumption}

\begin{Definition}[Sample distribution]\label{de:sample}
Under Assumption~\ref{as:select}, define the \emph{event of being selected} by $S = \{U \le \varphi\}$. The probability measure $Q$ on $(\Omega, \mathcal{F})$, defined by\vspace{12pt}
\begin{equation*}
Q[F] \ = \ P[F\,|\,S]\ = \ \frac{P[F\cap S]}{P[S]},\qquad \text{for}\ F \in \mathcal{F},
\end{equation*}
is called \emph{sample distribution}.
\end{Definition}
Note that the measure $Q$ is well-defined because from the independence of $U$ and $\bal{H}$, it follows that
\begin{equation*}
P[S] \ = \ E_P\left[\int_0^1 \mathbf{1}_{[0, \varphi]}(u)\, du \right] \ = \ E_P[\varphi] > 0.
\end{equation*}

Another consequence of the independence of $U$ and $\bal{H}$ is
\begin{equation}\label{eq:positive}
P[S\,|\,\bal{H}] \ = \ P[U \le \varphi\,|\,\bal{H}] \ = \ \varphi \ > \ 0.
\end{equation}

\begin{Proposition}\label{pr:satis}
\textls[-5]{$P$ and $Q$ as described in Assumption~\ref{as:select} and Definition~\ref{de:sample} satisfy
\mbox{Assumptions~\ref{as:setting} and \ref{as:cont}} with $P$ as source distribution and $Q$ as target distribution.
Moreover, $P$ is absolutely continuous with respect to $Q$ on $\bal{H}$.}
\end{Proposition}

\begin{proof}
It remains to show that
\begin{itemize}
\item $Q$ is absolutely continuous with respect to $P$ on $\bal{H}$, with density $\overline{h} = \frac{P[S\,|\,\bal{H}]}{P[S]}$;
\item $P$ is absolutely continuous with respect to $Q$ on $\bal{H}$;
\item $0 < Q[A_i]$ for $i = 1, \ldots, d$.
\end{itemize}

By definition of $Q$ as $P$ conditional on $S$, the sample distribution $Q$ is absolutely continuous with respect to $P$ on $\mathcal{F}$ and hence also on $\bal{H} \subset \mathcal{F}$.  For the density $\overline{h}$, we obtain
\begin{equation}\label{eq:hbar.sel}
\overline{h}\ = \ E_P\left[\frac{\mathbf{1}_S}{P[S]}\,\Big|\,\bal{H}\right]
\ = \ \frac{P[S\,|\,\bal{H}]}{P[S]} \ > \ 0.
\end{equation}

The fact that $\overline{h}$ is positive implies that $P$ is absolutely continuous with respect to $Q$ on $\bal{H}$. Since $A_i \in \bal{H}$ for $i= 1, \ldots, d$, the absolute continuity of $P$ with respect to $Q$  implies $Q[A_i] >0$, $i = 1, \ldots, d$.
\end{proof}

\subsection{Properties of the Sample Selection Model}
Equation~\eqref{eq:hbar.sel} implies for the density $h$ of $Q$ with respect to $P$ on $\mathcal{H}$
\begin{equation}\label{eq:h.sel}
h \ = \ E_P[\overline{h}\,|\,\mathcal{H}] \ = \ \frac{P[S\,|\,\mathcal{H}]}{P[S]} \ > \ 0.
\end{equation}

\textls[-15]{From representation \eqref{eq:Hbar} of $\bal{H}$, the following alternative description for $P[S\,|\,\bal{H}]$ follows:}
\begin{equation}\label{eq:choiceA}
\begin{split}
P[S\,|\,\bal{H}] & = \sum_{i=1}^d \frac{P[A_i\cap S\,|\,\mathcal{H}]}{P[A_i\,|\,\mathcal{H}]}\,\mathbf{1}_{A_i}\\
& = \sum_{i=1}^d P_i[S\,|\,\mathcal{H}]\,\mathbf{1}_{A_i},
\end{split}
\end{equation}
where the $P_i$ denote the class-conditional feature distributions under $P$, see Definition~\ref{de:condDist}.
$P_i[S\,|\,\mathcal{H}]$ is accordingly the feature-conditional probability of being selected on the subpopulation of objects with class $A_i$.

For $i = 1, \ldots, d$ and $H \in \mathcal{H}$, a short calculation shows:\vspace{12pt} 
\begin{align*}
Q_i[H] & = \frac{P[A_i \cap H \cap S]}{P[A_i \cap S]}\\
& = \frac{E_P\bigl[\mathbf{1}_H\,P[A_i \cap S\,|\,\mathcal{H}]\bigr]}{P[A_i \cap S]}\\
& = \frac{E_P\bigl[\mathbf{1}_H\,P[A_i \,|\,\mathcal{H}]\,P_i[S\,|\,\mathcal{H}]\bigr]}
{P[A_i]\,P_i[S]}\\
& = \frac{E_P\bigl[\mathbf{1}_{H\cap A_i}\,P_i[S\,|\,\mathcal{H}]\bigr]}
{P[A_i]\,P_i[S]}\\
& = E_{P_i}\left[\mathbf{1}_H\,\frac{P_i[S\,|\,\mathcal{H}]}{P_i[S]}\right].
\end{align*}

This implies
\begin{equation}\label{eq:hi.sel}
h_i\ =\ \frac{d Q_i}{d P_i}\Big|\mathcal{H}\ =\ \frac{P_i[S\,|\,\mathcal{H}]}{P_i[S]}, \quad
i = 1, \ldots, d.
\end{equation}

Equation~\eqref{eq:hi.sel} and Theorem~\ref{th:dens} together imply the following alternative representation of $\overline{h}$:
\begin{equation}\label{eq:normal.sel}
\overline{h}\ = \ \sum_{i=1}^d \frac{P_i[S\,|\,\mathcal{H}]}{P_i[S]}\,\frac{Q[A_i]}{P[A_i]}\,\mathbf{1}_{A_i}.
\end{equation}

By the generalised Bayes formula (Lemma~\ref{le:GenBayes} in Appendix~\ref{se:appendix}), \eqref{eq:hbar.sel} implies the following representation of the posterior class probabilities $Q[A_i\,|\,\mathcal{H}]$, $i = 1, \ldots, d$, under~$Q$:
\begin{align}
Q[A_i\,|\,\mathcal{H}] & =  \frac{E_P[\mathbf{1}_{A_i}\,\overline{h}\,|\,\mathcal{H}]}
{E_P[\overline{h}\,|\,\mathcal{H}]}\notag\\
& = \frac{E_P\bigl[\mathbf{1}_{A_i}\,P[S\,|\,\bal{H}]\,\big|\,\mathcal{H}\bigr]}
{P[S\,|\,\mathcal{H}]}\notag\\
& = \frac{P[S\cap A_i\,|\,\mathcal{H}]}{P[S\,|\,\mathcal{H}]}.\label{eq:Qsel}
\end{align}

Zadrozny \cite{Zadrozny2004selection} and Hein \cite{Hein2009binary} observed that if the event $S$ of being selected and the class labels as expressed by the $\sigma$-algebra $\mathcal{A}$ were independent conditional on $\mathcal{H}$, the information set reflecting the features, then the population distribution $P$ and the sample distribution $Q$ were related by covariate shift.
A consequence of \eqref{eq:Qsel} is that the converse of this observation actually also holds true, as stated in the following proposition.

\begin{Proposition}\label{pr:cov.sel}
In the sample selection model, as specified by Assumption~\ref{as:select} and Definition~\ref{de:sample}, the population distribution
$P$ and the sample distribution $Q$ are related by covariate shift if and only if
\begin{equation*}
P[S\cap A_i\,|\,\mathcal{H}] \ = \ P[S\,|\,\mathcal{H}]\,P[A_i\,|\,\mathcal{H}], \qquad i = 1, \ldots, d,
\end{equation*}
i.e.,\ if the event of being selected and the class labels are independent conditional on the features under the population distribution $P$.
\end{Proposition}
\begin{proof}
Proposition~\ref{pr:cov.sel} is obvious from \eqref{eq:Qsel} and the definition of covariate shift \eqref{eq:covariateShift}.
\end{proof}

In the case of general dataset shift caused by sample selection, Equation~\eqref{eq:Qsel} does not provide information about how to compute the population posterior class probabilities $P[A_i\,|\,\mathcal{H}]$ from the sample posterior class probabilities
$Q[A_i\,|\,\mathcal{H}]$.  Translated into the setting of this paper,  Hein \cite{Hein2009binary} presented in Equation~(3.2)  the following two ways to do so:
\begin{subequations}
\begin{itemize}
\item Define $Q^\ast$ as the distribution of the not-selected sample, i.e.,
\begin{equation*}
Q^\ast[F] \ = \ P[F\,|\,(\Omega\setminus S)] \ = \ \frac{P[F] - P[F\cap S]}{1-P[S]}, \qquad F\in \mathcal{F}.
\end{equation*}
Then, it holds that
\begin{equation}\label{eq:hein1}
\begin{split}
P[A_i\,|\,\mathcal{H}] & = P[A_i\cap S\,|\,\mathcal{H}] + P[A_i\cap (\Omega \setminus S)\,|\,\mathcal{H}]\\
& = Q[A_i\,|\,\mathcal{H}]\,P[S\,|\,\mathcal{H}]   + Q^\ast[A_i\,|\,\mathcal{H}]\,
(1-P[S\,|\,\mathcal{H}]),
\end{split}
\end{equation}
for $i=1, \ldots, d$.
\item Equation \eqref{eq:Qsel} can be written equivalently as
\begin{equation*}
Q[A_i\,|\,\mathcal{H}] \ = \ \frac{P_i[S\,|\,\mathcal{H}]\,P[A_i\,|\,\mathcal{H}]}{P[S\,|\,\mathcal{H}]}.
\end{equation*}
Hence, on the event $\bigl\{P_i[S\,|\,\mathcal{H}] > 0\bigr\}$, the following representation of $P[A_i\,|\,\mathcal{H}]$, $i = 1, \ldots, d$, is obtained:
\begin{equation}\label{eq:hein2}
P[A_i\,|\,\mathcal{H}] \ = \ \frac{P[S\,|\,\mathcal{H}]}{P_i[S\,|\,\mathcal{H}]}\,
Q[A_i\,|\,\mathcal{H}].
\end{equation}
\end{itemize}
\end{subequations}

Both \eqref{eq:hein1} and \eqref{eq:hein2} are of limited practical usefulness, however, as on the one hand, \eqref{eq:hein1} requires knowledge of the class labels in the not-selected sample, which usually are not available. On the other hand, for \eqref{eq:hein2} to be applicable, class-wise probabilities of selection $P_i[S\,|\,\mathcal{H}]$  must be estimated, which again requires knowledge of the class labels in the not-selected sample.

\subsection{Sample Selection Bias and Factorizable Joint Shift}

Proposition~\ref{pr:cov.sel} provides an example of a condition for the sample selection process that makes the resulting bias between population and sample representable as covariate shift and, consequently, according to Section~\ref{se:covshift}, as a special case of factorizable joint shift. Are there other selection procedures that entail factorizable joint shift?

We investigate this question by assuming that the population distribution $P$ and the sample distribution $Q$ are related by factorizable joint shift and then identifying the consequences this assumption implies for the class-wise feature-conditional selection probabilities $P_i[S\,|\,\mathcal{H}]$, $i=1,\ldots, d$.

\begin{Theorem}\label{th:FJS.sel}
Under Assumption~\ref{as:select} and Definition~\ref{de:sample}, let $P$ and $Q$ be related by factorizable joint shift in the sense of Definition~\ref{de:FJS}, i.e.,\ there are an $\mathcal{H}$-measurable function $g \ge 0$ and an $\mathcal{A}$-measurable function $b \ge 0$ such that the density $\overline{h}$ of $Q$ with respect to $P$ on $\bal{H}$ can be represented as $\overline{h} = g\,b$. Then, the following statements hold true:
\begin{itemize}
\item[(i)] $Q$ and $P$ are related by factorizable joint shift with an $\mathcal{H}$-measurable function $g^\ast > 0$ and an $\mathcal{A}$-measurable function $b^\ast >0$ that can be represented up to a constant factor in the sense of~\eqref{eq:nonunique}~as
\begin{subequations}
\begin{equation}\label{eq:rep.sel}
\begin{split}
b^\ast & = \sum_{i=1}^d \alpha_i\,\frac{P[A_i]}{Q[A_i]}\,\mathbf{1}_{A_i} +
\frac{P[A_d]}{Q[A_d]}\,\mathbf{1}_{A_d}\quad \text{and}\\
g^\ast & = \frac{P[S]}{P[S\,|\,\mathcal{H}]
\left(\sum_{i=1}^d \alpha_i\,\frac{P[A_i]}{Q[A_i]}\,Q[A_i\,|\,\mathcal{H}] +
\frac{P[A_d]}{Q[A_d]}\,Q[A_d\,|\,\mathcal{H}]\right)},
\end{split}
\end{equation}
where the constants $0 < \alpha_1, \ldots, \alpha_{d-1} < \infty$ satisfy the following equation system, with $i = 1, \ldots, d-1$:
\begin{equation}\label{eq:sys.sel}
Q[A_i] \ = \ P[S]\,\alpha_i\,E_Q\left[\frac{Q[A_i\,|\,\mathcal{H}]}{P[S\,|\,\mathcal{H}]
\left(\sum_{j=1}^d \alpha_j\,\frac{P[A_j]}{Q[A_j]}\,Q[A_j\,|\,\mathcal{H}] +
\frac{P[A_d]}{Q[A_d]}\,Q[A_d\,|\,\mathcal{H}]\right)} \right].
\end{equation}
\end{subequations}
\item[(ii)] The population posterior probabilities $P[A_i\,|\,\mathcal{H}]$, $i=1,\ldots,d$, can be represented as functions of the sample posterior probabilities $Q[A_i\,|\,\mathcal{H}]$, $i=1,\ldots,d$, in the following way:\vspace{12pt} 
\begin{equation}\label{eq:correct.sel}
\begin{split}
P[A_i\,|\,\mathcal{H}] & = \frac{\alpha_i\,\frac{P[A_i]}{Q[A_i]}\,Q[A_i\,|\,\mathcal{H}]}
{\sum_{j=1}^d \alpha_j\,\frac{P[A_j]}{Q[A_j]}\,Q[A_j\,|\,\mathcal{H}] +
\frac{P[A_d]}{Q[A_d]}\,Q[A_d\,|\,\mathcal{H}]}, \quad i=1,\ldots, d-1,\\
P[A_d\,|\,\mathcal{H}] & = \frac{\frac{P[A_d]}{Q[A_d]}\,Q[A_d\,|\,\mathcal{H}]}
{\sum_{j=1}^d \alpha_j\,\frac{P[A_j]}{Q[A_j]}\,Q[A_j\,|\,\mathcal{H}] +
\frac{P[A_d]}{Q[A_d]}\,Q[A_d\,|\,\mathcal{H}]},
\end{split}
\end{equation}
where the constants $0 < \alpha_1, \ldots, \alpha_{d-1} < \infty$ satisfy equation system \eqref{eq:sys.sel}.
\item[(iii)] The class-wise feature-conditional selection probabilities $P_i[S\,|\,\mathcal{H}]$, $i=1,\ldots,d$, can be represented as
\begin{equation}\label{eq:classwise}
P_i[S\,|\,\mathcal{H}]\ = \ \frac{Q[A_i]}{\alpha_i\,P[A_i]}\,P[S\,|\,\mathcal{H}]\,
\left(\sum_{j=1}^d \alpha_j\,\frac{P[A_j]}{Q[A_j]}\,Q[A_j\,|\,\mathcal{H}] +
\frac{P[A_d]}{Q[A_d]}\,Q[A_d\,|\,\mathcal{H}]\right),
\end{equation}
where the constants $0 < \alpha_1, \ldots, \alpha_{d-1} < \infty$ satisfy equation system
\eqref{eq:sys.sel} and $\alpha_d=1$.
\end{itemize}
\end{Theorem}

\begin{proof}
Functions $g$ and $b$ must be positive since $\overline{h}$ is positive according to Proposition~\ref{pr:satis}.
Hence, $Q$ and $P$ are related by factorizable joint shift with decomposition $b^\ast = 1/b$ and $g^\ast = 1/g$. Apply Theorem~\ref{th:factorized} with swapped roles of $P$ and $Q$ to obtain representation \eqref{eq:rep.sel} and equation system \eqref{eq:sys.sel}. Statement (ii) follows immediately from Corollary~\ref{co:factorized}.

Regarding (iii), use \eqref{eq:hi.sel} and Proposition~\ref{pr:reverse}~(iv) together with \eqref{eq:rep.sel} to obtain
\begin{equation*}
\frac{P_i[S]}{P_i[S\,|\,\mathcal{H}]}\ = \ \alpha_i\,g^\ast, \quad i=1,\ldots,d.
\end{equation*}

This is equivalent to \eqref{eq:classwise}.
\end{proof}


As mentioned  in Section~\ref{se:alternative} as a potential application of Theorem~\ref{th:factorized}, assuming that the posterior probabilities $Q[A_i\,|\,\mathcal{H}]$ under the sample distribution can be estimated, Theorem~\ref{th:FJS.sel} offers two obvious ways to learn the characteristics of factorizable joint shift:
\begin{itemize}
\item[(a)] If the population prior class probabilities $P[A_i]$ are known (for instance from external sources) solve \eqref{eq:sys.sel} for the constants $\alpha_i$.
\item[(b)] If the population prior class probabilities $P[A_i]$ are unknown, fix values for the constants $\alpha_i$ and solve \eqref{eq:sys.sel} for the $P[A_i]$. Letting $\alpha_i =1$ for all $i$ is a natural choice that converts \eqref{eq:sys.sel} into the system of maximum likelihood equations for the $P[A_i]$ under the prior probability shift assumption.
\end{itemize}

In case (a), \eqref{eq:classwise} may serve as an admissibility check for the solutions found. If the class-wise selection probabilities $P_i[S\,|\,\mathcal{H}]$ obtained from \eqref{eq:classwise} can take values greater than $100\%$, the corresponding set of values $(\alpha_1, \ldots, \alpha_{d-1})$ is not an admissible solution of \eqref{eq:sys.sel}. If all solutions $(\alpha_1, \ldots, \alpha_{d-1})$ of \eqref{eq:sys.sel} turn out to be inadmissible, it must be concluded that the assumption of factorizable joint shift for the sample selection process is wrong.

In case (b), from \eqref{eq:classwise} follows for all $i, j =1, \ldots, d$
\begin{equation*}
P_i[S\,|\,\mathcal{H}]\,\frac{P[A_i]}{Q[A_i]} \ =\  P_j[S\,|\,\mathcal{H}]\,\frac{P[A_j]}{Q[A_j]},
\end{equation*}
which implies
\begin{equation}
P_i[S\,|\,\mathcal{H}] \ \le\ \frac{Q[A_i]}{P[A_i]}\,
\min\left(\frac{P[A_1]}{Q[A_1]}, \ldots, \frac{P[A_d]}{Q[A_d]}\right), \quad
\text{for all}\ i = 1, \ldots, d.\label{eq:necessary}
\end{equation}

Inequality~\eqref{eq:necessary} provides a simple necessary criterion for the presence of factorizable joint shift with constants $\alpha_i$ all equal to 1.

A further, less obvious special case of Theorem~\ref{th:FJS.sel} is encountered if is assumed that\vspace{12pt} 
\begin{equation}
\alpha_i\,\frac{P[A_i]}{Q[A_i]}\ = \ \frac{P[A_d]}{Q[A_d]}, \qquad \text{for all}\ i=1, \ldots, d-1.
\end{equation}

Then, \eqref{eq:classwise} implies $P_i[S\,|\,\mathcal{H}] = P[S\,|\,\mathcal{H}]$ for all $i=1,\ldots, d$.
By \eqref{eq:hein2}, this means that population distribution and sample distribution are related by covariate shift, as already observed by Hein \cite{Hein2009binary}.

\section{Conclusions}
\label{se:concl}

\textls[-15]{We revisited the notion of ``factorizable joint shift'' recently introduced by \mbox{He et al.~\cite{he2022domain}}. 
A main finding is that factorizable joint shift is actually not much more general than prior probability shift or covariate shift. However, in contrast to these two types of shifts, factorizable joint shift is not fully identifiable if no class label information
on the test (target) dataset is available and no additional assumptions are made.
These findings are based on a representation result (Theorem~\ref{th:factorized}) and a comparison of the class posterior correction formula \eqref{eq:FJScorrect} for factorizable joint shift to the related correction formulae \eqref{eq:priorcorrect}
and \eqref{eq:covariateShift} for prior probability and covariate shifts, respectively. Formula \eqref{eq:FJScorrect} is structurally identical with formula \eqref{eq:priorcorrect} but includes additional constants which can be found by solving the nonlinear equation system~\eqref{eq:system}.}

\textls[-10]{He et al.~\cite{he2022domain} did not present the full rationale for their joint importance aligning approach to estimating the characteristics of factorizable joint shift. Hence, solving equation system~\eqref{eq:system} for the additional constants in the posterior correction formula or for the prior class probabilities under the target distribution can be considered attractive alternative~approaches.}

Some open research questions remain:
\begin{itemize}
\item Under what conditions can the existence and the uniqueness of solutions $(\varrho_1, \ldots, \varrho_{d-1})$ to equation system \eqref{eq:system} be guaranteed in the case of more than two classes?
\item Is there any manageable---in the sense of having observable characteristics---type of
dataset shift which is both more complex than factorizable joint shift and less complex than
covariate shift with posterior drift?
\item To which extent can Theorem~\ref{th:factorized} be adapted for a more general regression setting?
\end{itemize}
\vspace{6pt}

\funding{This research received no external funding.}

\institutionalreview{\hl{Not applicable.}}

\informedconsent{\hl{Not applicable.}}

\dataavailability{\hl{Not applicable.}} 

\acknowledgments{The author thanks three anonymous reviewers for suggestions that helped to
improve an earlier version of this paper.}

\conflictsofinterest{The author declares no conflict of interest.}

\appendixtitles{yes} 
\appendixstart
\appendix
\section[\appendixname~\thesection]{The Generalized Bayes Formula}
\label{se:appendix}

Lemma~\ref{le:GenBayes} is Theorem~10.8 of Klebaner \cite{Klebaner}, slightly extended to explicitly cover the case when the denominator in the formula for the density can be $0$.

\begin{Lemma}\label{le:GenBayes}
Let $(\Omega, \mathcal{F})$ be a measurable space and $P$ and $Q$ probability measures on $(\Omega, \mathcal{F})$. Assume that $f = \frac{d Q}{d P}$ is a density of $Q$ with respect to $P$ on $\mathcal{F}$.
Let $\mathcal{G}$ be a sub-$\sigma$-algebra of $\mathcal{F}$ and $X$ be a non-negative random variable on $(\Omega, \mathcal{F})$ or
a random variable on $(\Omega, \mathcal{F})$ such that $f\,X$ is $P$-integrable.
Then, the following two statements hold:
\begin{itemize}
\item[(i)] $\{f>0\} \subset \bigl\{E_P[f\,|\,\mathcal{G}] > 0\bigr\}$, in the sense of $P\bigl[f> 0,\,E_P[f\,|\,\mathcal{G}] = 0\bigr] =0$.
\item[(ii)] $\displaystyle{}E_Q[X\,|\,\mathcal{G}] \ = \
\frac{E_P[f\,X\,|\,\mathcal{G}]}{E_P[f\,|\,\mathcal{G}]}
\mathbf{1}_{\{E_P[f\,|\,\mathcal{G}] > 0\}}$.
\end{itemize}
\end{Lemma}
\begin{proof}
\noindent{}For (i): Observe that
\begin{equation*}
E_P[f\,\mathbf{1}_{\{E_P[f\,|\,\mathcal{G}] = 0\}}] \ = \
E_P\bigl[E_P[f\,|\,\mathcal{G}]\,\mathbf{1}_{\{E_P[f\,|\,\mathcal{G}] = 0\}}\bigr] \ = \ 0.
\end{equation*}

This implies
\begin{equation*}
0 \ = \ P\bigl[f\,\mathbf{1}_{\{E_P[f\,|\,\mathcal{G}] = 0\}} > 0\bigr]
\ = \ P\bigl[f> 0,\,E_P[f\,|\,\mathcal{G}] = 0\bigr].
\end{equation*}

For (ii): see Klebaner \cite{Klebaner}, proof of Theorem~10.8.
\end{proof}

\section[\appendixname~\thesection]{Proofs}
\label{se:proofs}

\subsection{Proof of Proposition~\ref{pr:d2}}

For a more concise notation, define the non-negative, $\mathcal{H}$-measurable random variables $R_1$ and $R_2$ by
\begin{equation*}
R_1\ = \ \frac{P[A\,|\,\mathcal{H}]}{p}\quad \text{and}\quad R_2 \ = \frac{1-P[A\,|\,\mathcal{H}]}{1-p.}
\end{equation*}
\begin{subequations}

Then, \eqref{eq:system} can be written as
\begin{equation}\label{eq:orig}
1 \ = \ \varrho\,E_P\left[\frac{h\,R_1}{\varrho\,q\,R_1+(1-q)\,R_2}\right].
\end{equation}

Some algebra shows that \eqref{eq:orig} is equivalent to
\begin{equation}\label{eq:equiv1}
1 \ = \ E_P\left[\frac{h\,R_2}{\varrho\,q\,R_1+(1-q)\,R_2}\right],
\end{equation}
and that it is also equivalent to
\begin{equation}\label{eq:equiv2}
0 \ = \ E_P\left[h \frac{\varrho\,R_1 - R_2}{\varrho\,q\,R_1+(1-q)\,R_2}\right].
\end{equation}
\end{subequations}

Define the function $g(\varrho) = E_P\left[\frac{h\,R_2}{\varrho\,q\,R_1+(1-q)\,R_2}\right]$ for $\varrho \ge 0$. Then, it holds that
\begin{itemize}
\item $g(\varrho) \le \frac{1}{1-q} < \infty$ for all $\varrho \ge 0$;
\item $g(0) = \frac{1}{1-q} > 1$;
\item By the dominated convergence theorem, $g$ is continuous for $0 \le \varrho < \infty$ with $\lim\limits_{\varrho\to\infty} g(\varrho) = 0$.
\end{itemize}

By the mean value theorem, these properties of $g$ imply the existence of some $\varrho > 0$ with $g(\varrho) = 1$. By the equivalence of \eqref{eq:orig} and \eqref{eq:equiv1}, the existence of a positive solution $\varrho$ to \eqref{eq:system} follows.

Regarding the uniqueness of the solution to \eqref{eq:system}, define for $q \in (0,1)$ and $\varrho \in (0, \infty)$ the function $f(q, \varrho) = E_P\left[h \frac{\varrho\,R_1 - R_2}{\varrho\,q\,R_1+(1-q)\,R_2}\right]$. Then, $f$ is continuously partially differentiable~with
\begin{align*}
\frac{\partial f}{\partial q}(q, \varrho) & = - E_P\left[
h \frac{(\varrho\,R_1 - R_2)^2}{(\varrho\,q\,R_1+(1-q)\,R_2)^2}\right], \quad \text{and}\\
\frac{\partial f}{\partial \varrho}(q, \varrho) & = E_P\left[
h \frac{R_1\,R_2}{(\varrho\,q\,R_1+(1-q)\,R_2)^2}\right].
\end{align*}

The assumption $P\bigl[P[A\,|\,\mathcal{H}]\in\{0,1\}\bigr] = 0$ implies $\frac{\partial f}{\partial \varrho}(q, \varrho) >0$ for all $0<q<1$ and $\varrho > 0$.

$P[\varrho\,R_1 - R_2 = 0] =1$ would imply $P[A\,|\,\mathcal{H}] = p$ and as a further consequence $A$ and $\mathcal{H}$ would be independent. By assumption, this is not the case, and hence, $P[\varrho\,R_1 - R_2 = 0] < 1$. This implies also $\frac{\partial f}{\partial q}(q, \varrho) > 0$ for all $0<q<1$ and $\varrho > 0$.

Consequently, by the implicit function theorem, there exists a continuously differentiable function $\phi:(0,1) \to (0,\infty)$, $q \mapsto \phi(q)=\varrho$ such that $f(q, \phi(q)) = 0$ for all $0 < q < 1$~and\vspace{12pt} 
\begin{equation*}
\phi'(q) \ = \ \frac{E_P\left[
h \frac{(\phi(q)\,R_1 - R_2)^2}{(\phi(q)\,q\,R_1+(1-q)\,R_2)^2}\right]}
{E_P\left[
h \frac{R_1\,R_2}{(\phi(q)\,q\,R_1+(1-q)\,R_2)^2}\right]}\ > \ 0.
\end{equation*}

This proves claim (i) on $\phi$ and the existence of $\lim_{q\to 0} \phi(q) < \infty$ and $\lim_{q\to 1} \phi(q) > 0$.
Making use again of the equivalence of \eqref{eq:equiv2} and \eqref{eq:orig} and invoking 
Lemma~4.1 of Tasche~\cite{tasche2013law} now implies for all $0 < q < 1$
\begin{equation}\label{eq:ineq}
\frac{1}{E_P\left[h \frac{R_1}{R_2} \right]} \ < \ \phi(q) \ < \ E_P\left[h \frac{R_2}{R_1}\right].
\end{equation}

These inequalities also hold true if $E_P\left[h \frac{R_1}{R_2} \right] = \infty$ or $E_P\left[h \frac{R_2}{R_1}\right] = \infty$.

Now, apply Fatou's lemma to obtain
\begin{align*}
1 & = \liminf\limits_{q\to 1} E_P\left[h \frac{R_2}{\phi(q)\,q\,R_1 + (1-q)\,R_2}\right]\\
& \ge  E_P\left[h \frac{R_2}{R_1} \frac{1}{\lim\limits_{q\to 1} \phi(q)}\right].
\end{align*}
From this, $\lim_{q\to 1} \phi(q) \ge E_P\left[h \frac{R_2}{R_1}\right]$ follows, and by \eqref{eq:ineq}, also claim (iii).

Another application of Fatou's lemma gives
\begin{align*}
1 & = \liminf\limits_{q\to 0} \left(\phi(q)\,E_P\left[h \frac{R_1}{\phi(q)\,q\,R_1 +
(1-q)\,R_2}\right]\right)\\
& \ge   E_P\left[h \frac{R_1}{R_2} \lim\limits_{q\to 1} \phi(q)\right].
\end{align*}

Together with \eqref{eq:ineq}, this proves claim (ii) and completes the proof.

\subsection{Proof of Proposition~\ref{pr:GLS}}

Observe that function $g$ in \eqref{eq:GLSsplit} is well-defined because the denominator
\begin{equation}\label{eq:gamma}
\gamma = \sum_{i=1}^d \frac{Q[A_i]}{P[A_i]} P[A_i\,|\,\mathcal{G}] =
\sum_{i=1}^d \frac{Q[A_i]}{P[A_i]} P[A_i\,|\,\mathcal{H}]
\end{equation}
on the right-hand side of the equation is always positive.
\begin{subequations}
On the one hand, by Corollary~\ref{co:correct}, we obtain for $i = 1, \ldots, d$ on the set $\{h >0\}$
\begin{equation}\label{eq:full}
Q[A_i\,|\,\mathcal{H}] \ = \ h_i\, \frac{Q[A_i]}{P[A_i]}\, \frac{P[A_i\,|\,\mathcal{H}]}{h},
\end{equation}
where $h_i$ denotes the density of the target class-conditional feature distribution $Q_i$ with respect to the source class-conditional feature distribution $P_i$ on $\mathcal{H}$.

On the other hand, by combining the prior probability shift property \eqref{eq:GLS} on $\mathcal{G}=\sigma(T)$, the sufficiency property \eqref{eq:sufficiency} and \eqref{eq:priorhi}, Corollary~\ref{co:correct} implies
\begin{equation}\label{eq:reduc}
Q[A_i\,|\,\mathcal{H}] \ = \ \frac{Q[A_i]}{P[A_i]}\, \frac{P[A_i\,|\,\mathcal{H}]}{\gamma},
\end{equation}

Hence, from \eqref{eq:full} and \eqref{eq:reduc}, it follows for $i = 1, \ldots, d$
\begin{equation}\label{eq:aux}
\frac{P[A_i\,|\,\mathcal{H}]}{\gamma} \ = \
h_i\, \frac{P[A_i\,|\,\mathcal{H}]}{h}\quad \text{on}\ \{h > 0\}.
\end{equation}
\end{subequations}

Making use of \eqref{eq:aux}, we obtain for any $F = \bigcup_{i=1}^d (A_i \cap H_i) \in \bal{H}$
\begin{align*}
Q[F] & = E_Q\bigl[Q[F\,|\,\mathcal{F}]\bigr] \\
& = \sum_{i=1}^d E_P\bigl[h\,\mathbf{1}_{H_i}\,Q[A_i\,|\,\mathcal{H}]\bigr]\\
& = \sum_{i=1}^d E_P\left[h\,\mathbf{1}_{\{h>0\}}\,\mathbf{1}_{H_i}\,
h_i\, \frac{Q[A_i]}{P[A_i]}\, \frac{P[A_i\,|\,\mathcal{H}]}{h}\right]\\
& = \sum_{i=1}^d E_P\left[h\,\mathbf{1}_{H_i}\,
\frac{Q[A_i]}{P[A_i]}\, \frac{P[A_i\,|\,\mathcal{H}]}{\gamma}\right]\\
& = E_P\left[\frac{h}{\gamma} \sum_{i=1}^d \frac{Q[A_i]}{P[A_i]}\,
\mathbf{1}_{H_i}\,P[A_i\,|\,\mathcal{H}]\right]\\
& = E_P\left[\frac{h}{\gamma} \sum_{i=1}^d \frac{Q[A_i]}{P[A_i]}\,
\mathbf{1}_{A_i\cap H_i}\right]\\
& = E_P\left[\mathbf{1}_F\,\frac{h}{\gamma} \sum_{i=1}^d \frac{Q[A_i]}{P[A_i]}\,
\mathbf{1}_{A_i}\right].
\end{align*}

This proves that $b\,g$ with $b$ and $g$ as defined by \eqref{eq:GLSsplit} is a density of $Q$ with respect to $P$ on $\bal{H}$. As the $\mathcal{A}$-measurability of $b$ and $\mathcal{H}$-measurability of $g$ are obvious, the proof is~complete.

\begin{adjustwidth}{-\extralength}{0cm}
\reftitle{References}




%
%
%
\end{adjustwidth}
\end{document}